\documentclass[runningheads]{llncs}

\usepackage{eccv}

\usepackage{eccvabbrv}
\usepackage{pifont}
\usepackage{adjustbox}
\usepackage{subcaption}
\usepackage{hhline}
\usepackage{diagbox}
\usepackage{graphicx}
\usepackage{booktabs}

\usepackage{multirow}
\usepackage[accsupp]{axessibility}  

\usepackage{blindtext}
\usepackage{color}
\usepackage{bbm, dsfont}

\usepackage{colortbl}  
\usepackage{xcolor}
\usepackage{array}

\usepackage{hyperref}

\usepackage{orcidlink}

\begin{document}

\title{RAW-Adapter: Adapting Pre-trained Visual Model to Camera RAW Images} 

\titlerunning{RAW-Adapter}

\author{Ziteng Cui\inst{1}~\thanks{corresponding author} \and
Tatsuya Harada\inst{1,2}}

\authorrunning{Cui, Harada.}

\institute{The University of Tokyo \and
RIKEN AIP\\
\email{\{cui, harada\}@mi.t.u-tokyo.ac.jp}}


\maketitle

\begin{abstract}
sRGB images are now the predominant choice for pre-training visual models in computer vision research, owing to their ease of acquisition and efficient storage. Meanwhile, the advantage of RAW images lies in their rich physical information under variable real-world challenging lighting conditions. For computer vision tasks directly based on camera RAW data, most existing studies adopt methods of integrating image signal processor (ISP) with backend networks, yet often overlook the interaction capabilities between the ISP stages and subsequent networks. 
Drawing inspiration from ongoing adapter research in NLP and CV areas, we introduce \textbf{RAW-Adapter}, a novel approach aimed at adapting sRGB pre-trained models to camera RAW data. RAW-Adapter comprises input-level adapters that employ learnable ISP stages to adjust RAW inputs, as well as model-level adapters to build connections between ISP stages and subsequent high-level networks.
Additionally, RAW-Adapter is a general framework that could be used in various computer vision frameworks. Abundant experiments under different lighting conditions have shown our algorithm's state-of-the-art (SOTA) performance, demonstrating its effectiveness and efficiency across a range of real-world and synthetic datasets. Code is available at \href{https://github.com/cuiziteng/ECCV_RAW_Adapter}{this url}.
\end{abstract}

\section{Introduction}
\label{sec:intro}

In recent years, there has been a growing interest in revisiting vision tasks using  unprocessed camera RAW images. How to leverage information-rich RAW image in computer vision tasks has become a current research hotspot in various sub-areas (\textit{i.e.} denoising~\cite{brooks2019unprocessing,zamir2020cycleisp}, view synthesis~\cite{RAW_NeRF}, object detection~\cite{Hardware_in_the_loop,LOD_BMVC2021}). Compared to commonly used sRGB image, RAW image directly acquired by the camera sensor, encompasses abundant information unaffected or compressed by image signal processor (ISP), also offers physically meaningful information like noise distributions~\cite{Dancing_under_light,Wei_2020_CVPR}, owing to its linear correlation between image intensity and the radiant energy received by a camera. The acquisition of RAW data, facilitating enhanced detail capture and a higher dynamic range, imparts a unique advantage in addressing visual tasks under variable lighting conditions in the real world. For instance, sunlight irradiance can reach as high as $1.3 \times 10^3 W/m^2$, while bright planet irradiance can be as low as $2.0 \times 10^{-6} W/m^2$~\cite{Siggraph_Physic_night_sky}.

\begin{figure*}[t]
    \centering
    \includegraphics[width=1.00\linewidth]{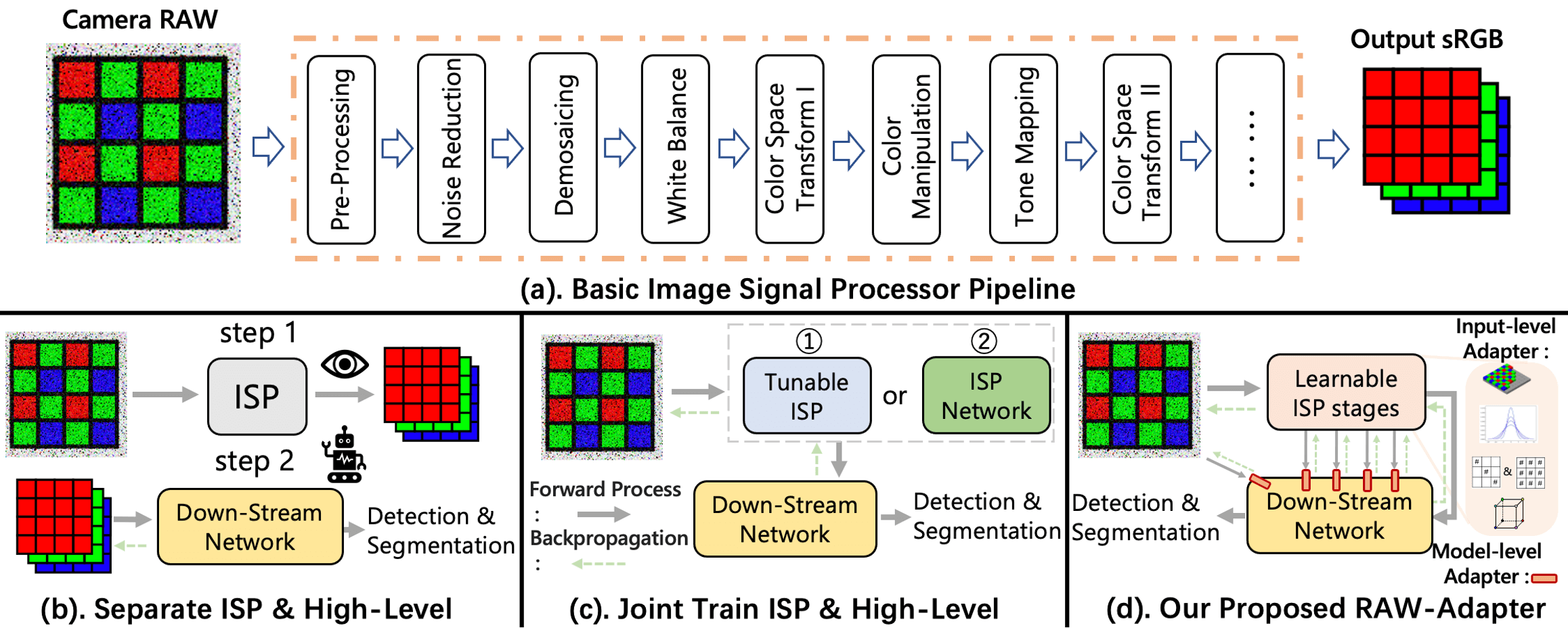}
    \caption{(a). An overview of basic image signal processor (ISP) pipeline. (b). ISP and current visual model have different objectives. (c) Previous methods optimize ISP with down-stream visual model. (d) Our proposed RAW-Adapter.}
    \label{fig:motivation}
\end{figure*}

 Meanwhile, sRGB has emerged as the primary choice for pre-training visual models in today's computer vision field, due to its scalability and ease of storage. Typically, sRGB images  are derived from camera RAW data through the ISP pipeline. As shown in Fig.~\ref{fig:motivation}(a), the entire ISP pipeline consists of multiple modules to convert RAW images into vision-oriented sRGB images, each of these modules serves its own distinct purpose, with the majority being ill-posed and heavily dependent on prior information~\cite{ISP_2005,Michael_eccv16,heide2014flexisp,Camera_Net,Mobile_Computational,Joint_denoising_demosaicing}. 

  When adopt RAW data for computer vision tasks, the purpose of a manually designed ISP is to produce images that offer a superior visual experience~\cite{Vision_ISP_ICIP,steven:dirtypixels2021}, rather than optimizing for downstream visual tasks such as object detection or semantic segmentation (see  Fig.~\ref{fig:motivation}(b)). Additionally, most companies' ISPs are black boxes, making it difficult to obtain information about the specific steps inside. Consequently, utilizing human vision-oriented ISP in certain conditions is sometimes even less satisfactory than directly using RAW~\cite{Bayer_RAW_HOG,LOD_BMVC2021,RAW_segment_dataset,Guo_2024_CVPR}.

\begin{figure*}[t]
    \centering
    \includegraphics[width=1.00\linewidth]{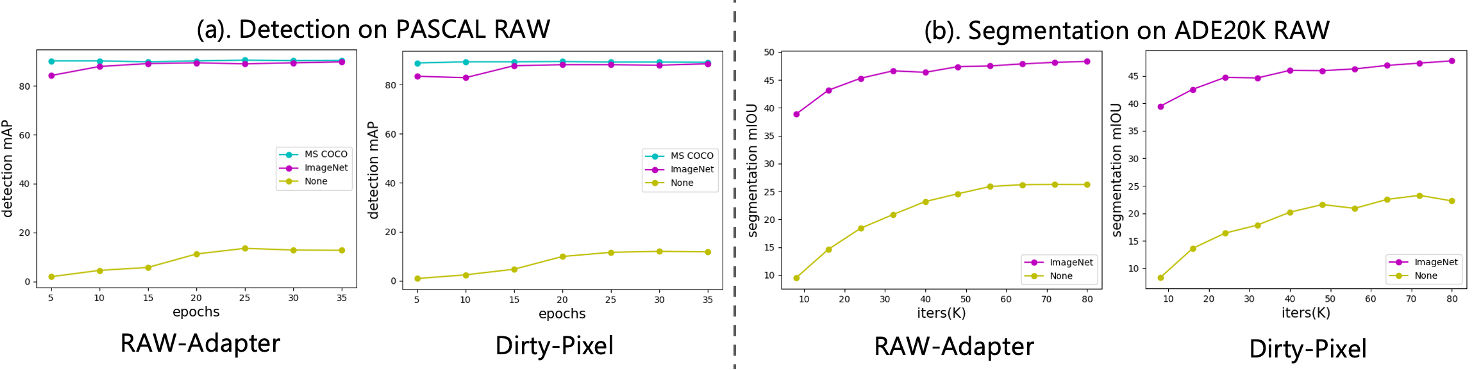}
    \caption{Performance of RAW-based visual tasks with and without sRGB pre-trained weights. We analyze two methods: Dirty-Pixel~\cite{steven:dirtypixels2021} and RAW-Adapter. Blue line represents trained with MS COCO~\cite{COCO_dataset} pre-train weights, the purple line indicates ImageNet~\cite{imagenet_cvpr09} pre-train weights, and the yellow line signifies training from scratch.}
    \label{fig:pretrain_weights}
    
\end{figure*}

To better take advantage of camera RAW data for various computer vision tasks. Researchers began to optimize the image signal processor (ISP) jointly with downstream networks. Since most ISP stages are non-differentiable and cannot be jointly backpropagation, there are two main lines of approaches to connect ISP stages with downstream networks (see Fig.~\ref{fig:motivation}(c)): \ding{172}. First-kind approaches maintain the modular design of the traditional ISP, involving the design of differentiable ISP modules through optimization algorithms~\cite{Hardware_in_the_loop,Vision_ISP_ICIP,yu2021reconfigisp}, such as Hardware-in-the-loop~\cite{Hardware_in_the_loop} adopt covariance matrix adaptation evolution strategy (CMA-ES)~\cite{CMA-ES}. \ding{173}. Second-kind approaches replaced the ISP part entirely with a neural network~\cite{Attention_ISP_ECCV,steven:dirtypixels2021,DynamicISP_2023_ICCV,RAW_OD_CVPR2023}, such as Dirty-Pixel~\cite{steven:dirtypixels2021} adopt a stack of residual UNets~\cite{UNet} as pre-encoder, however, the additional neural network introduces a significant computational burden, especially when dealing with high-resolution inputs. 
Beyond these, the above-mentioned methods still treat ISP and the backend neural network as two independent modules, lacking interaction capabilities between two separate models. Especially, current visual models predominantly pre-trained on large amounts of sRGB images, 
there exists a significant gap between pre-trained models and camera RAW images. As it shown in Fig.~\ref{fig:pretrain_weights}, training from scratch with RAW data can significantly impair the performance of RAW-based vision tasks. How to leveraging sRGB pre-trained weights proves to be highly beneficial for RAW image visual tasks.
Hence, here we ask: Is there a more effective way to adapt the \textbf{information-rich} RAW data into the \textbf{knowledge-rich} sRGB pre-trained models?

Our solution, RAW-Adapter, differs from previous approaches that employed complex ISP or deeper neural networks at the input level. Instead, we prioritize simplifying the input stages and enhancing connectivity between ISP and subsequent networks at the model level. Inspired by recent advancements in prompt learning and adapter tuning~\cite{Prompt_NLP,VPT_ECCV2022,Vit_adapter,potlapalli2023promptir}, we developed two novel adapter approaches aimed at enhancing the integration between RAW input and sRGB pre-trained models. Leveraging priors from ISP stages, our method includes input-level adapters and model-level adapters, as illustrated in Fig.~\ref{fig:motivation}(d). Input-level adapters are designed to adapt the input from RAW data to the backend network input, we follow traditional ISP design and predict ISP stages' key parameters to accommodate RAW image into the specific down-stream vision tasks, in this part, we employ strategies involving Query Adaptive Learning (QAL) and Implicit Neural Representation (INR) to ensure the differentiability of the core ISP processes while also upholding a lightweight design. Model-level adapters leverage prior knowledge from the input-level adapters by extracting intermediate stages as features, these features are then embedded into the downstream network, thus the network incorporates prior knowledge from the ISP stages, then jointly contributes to final machine-vision perception tasks.
Our contributions could be summarized as follows:

\begin{itemize}
    \item We introduce RAW-Adapter, a novel framework aimed at improving the integration of sRGB pre-trained models to camera RAW images. Through the implementation of two types of adapters, we strive to narrow the disparity between RAW images and sRGB models, tackling discrepancies at both input and model levels.
    
    \item By analyzing camera ISP steps, we've crafted input-level adapters that integrate query adaptive learning (QAL) and implicit neural representations (INR) to optimize ISP key parameters. Additionally, harnessing prior input stage information informs model-level adapters, enriching model understanding and improving downstream task performance.

    \item We conducted detection and segmentation experiments involving different lighting scenarios, including normal-light, dark, and over-exposure scenes. Through comparisons with current mainstream ISPs and joint-training methods, sufficient experiments demonstrate that our algorithm achieved state-of-the-art (SOTA) performance.
\end{itemize}

\section{Related Works}
\label{sec:related_work}
\subsection{Image Signal Processor}

Typically, a camera's Image Signal Processor (ISP) pipeline is required to reconstruct a high-quality sRGB image from camera RAW data, traditional ISP pipeline is formulated as a series of manually crafted modules executed sequentially~\cite{ISP_2005,Michael_eccv16,Mobile_Computational,ICIP_Parameter_Tuning}, including some representative steps such as demosaicing, white balance, noise removal, tone mapping, color space transform and so on. There also exist alternative designs of ISP process such as Heide~\textit{et al.}~\cite{heide2014flexisp} designed FlexISP which combine numerous ISP blocks to a joint optimization block, and Hasinoff~\textit{et al.}~\cite{Siggraph16_low_light_ISP} modified some of the traditional ISP steps for burst photography under extreme low-light conditions. When it comes to deep learning era, Chen~\textit{et al.}~\cite{SID} proposed SID to use a UNet~\cite{UNet} replace the traditional ISP steps, which translate low-light RAW data to normal-light sRGB images, and Hu~\textit{et al.}~\cite{hu_white_box} proposed a white-box solution with eight differentiable filters. After that, many efforts began to substitute the traditional ISP process with a neural network, such as~\cite{Deep_ISP,Camera_Net,PyNet,RAW-to-sRGB_ICCV2021,Meta_ISP,AAAI_learnable_ISP,SID,jincvpr23dnf,kim2024paramisp}, however deep network models are constrained by the training dataset, leading to shortcomings in generalization performance. 
Meanwhile another line of work focus on translate sRGB images back to RAW data~\cite{reverse_ISP_ICIP,64KB_RGB2RAW,Nam2017ModellingTS,Wang_2023_CVPR,brooks2019unprocessing,invertible_ISP,zamir2020cycleisp}. Apart from the conversion between RAW and RGB, there is ongoing research specifically dedicated to harnessing RAW images for downstream computer vision tasks. 

\subsection{Computer Vision based on RAW data}

In order to effectively leverage information in RAW images for downstream visual tasks and save the time required for ISP, some early methods proposed to directly perform computer vision tasks on RAW images~\cite{buckler2017reconfiguring,Bayer_RAW_HOG}, which lacks consideration for the camera noise introduced during the conversion from photons to RAW images, especially in low-light conditions~\cite{Wei_2020_CVPR,Dancing_under_light,zhuoxiao_Li}. Meanwhile training from scratch on RAW data would abandon the current large-scale pre-trained visual model on sRGB data (see Fig.~\ref{fig:pretrain_weights}), especially the quantity of RAW image datasets (\textit{i.e.} 4259 images in PASCAL RAW~\cite{omid2014pascalraw} dataset and 2230 images in LOD~\cite{LOD_BMVC2021} dataset) is incomparable to current RGB datasets (\textit{i.e.} over 1M images in ImageNet~\cite{imagenet_cvpr09} dataset and SAM~\cite{Segment_Anything} is trained with 11M images). 

Most subsequent research focuses on finding ways to integrate ISP and backend computer vision models~\cite{Hardware_in_the_loop,steven:dirtypixels2021,RAW_OD_CVPR2023,Vision_ISP_ICIP,yu2021reconfigisp,Attention_ISP_ECCV,DynamicISP_2023_ICCV,Guo_2024_CVPR}. For example Wu~\textit{et al.}~\cite{Vision_ISP_ICIP} first propose VisionISP and  emphasized the difference between human vision and machine vision, they introduced several trainable modules in the ISP to enhance backend object detection performance. Since then several methods attempt to replace the existing non-differentiable ISP with a differentiable ISP network~\cite{Hardware_in_the_loop,yu2021reconfigisp} or directly using encoder networks to replace the ISP process~\cite{steven:dirtypixels2021,Attention_ISP_ECCV,DynamicISP_2023_ICCV}, however training two consecutive networks simultaneously can result in a significant consumption of computational resources,previous work also rarely focused on how to fine-tune the current mainstream sRGB visual models more efficiently on RAW data. Additionally, research like Rawgment~\cite{Rawgment_2023_CVPR}, focusing on achieving realistic data augmentation directly on RAW images, some studies address the decrease in computer vision performance caused by specific steps within the in-camera process or ISP, such as white balance error~\cite{Error_WB_Vision}, auto-exposure error~\cite{Exposure_CVPR2021}, camera motion blur~\cite{Blurry_detection} and undesirable camera noise~\cite{goyal2022robust,cui2022exploring}.

\subsection{Adapters in Computer Vision}

Adapters have become prevalent NLP area, which introducing new modules in large language model (LLM) such as~\cite{Prompt_NLP,stickland2019bert}, these modules enable task-specific fine-tuning, allowing pre-trained models to swiftly adapt to downstream NLP tasks. In the computer vision area, adapters have been adopted in various areas such as incremental learning~\cite{iscen2020memory} and domain adaptation~\cite{adapter_DA}. Recently, a series of adapters have been used to investigate how to better utilize pre-trained visual models~\cite{Vit_adapter,VPT_ECCV2022,sam_hq} or pre-trained vision-language models~\cite{sung2022vl,zhang2021tip}. These methods focus on utilizing prior knowledge to make pre-trained models quickly adapt to downstream tasks. Here, we introduce the RAW-Adapter to further improve the alignment between RAW inputs and sRGB pre-trained visual models.

\section{RAW-Adapter}

In this section, we introduce RAW-Adapter. 
Due to page limitation, an overview of the conventional image signal processor (ISP) in supplementary's Sec.~\ref{sec3.1ISP_revisited}. For specific adapter designs, we selectively omit certain steps and focus on some key steps within the camera ISP. In Sec.~\ref{sec3.2input}, we introduce RAW-Adapter's input-level adapters, followed by an explanation of its model-level adapters in Sec.~\ref{sec3.3model}.

\begin{figure*}[t]
    \centering
    \includegraphics[width=1.00\linewidth]{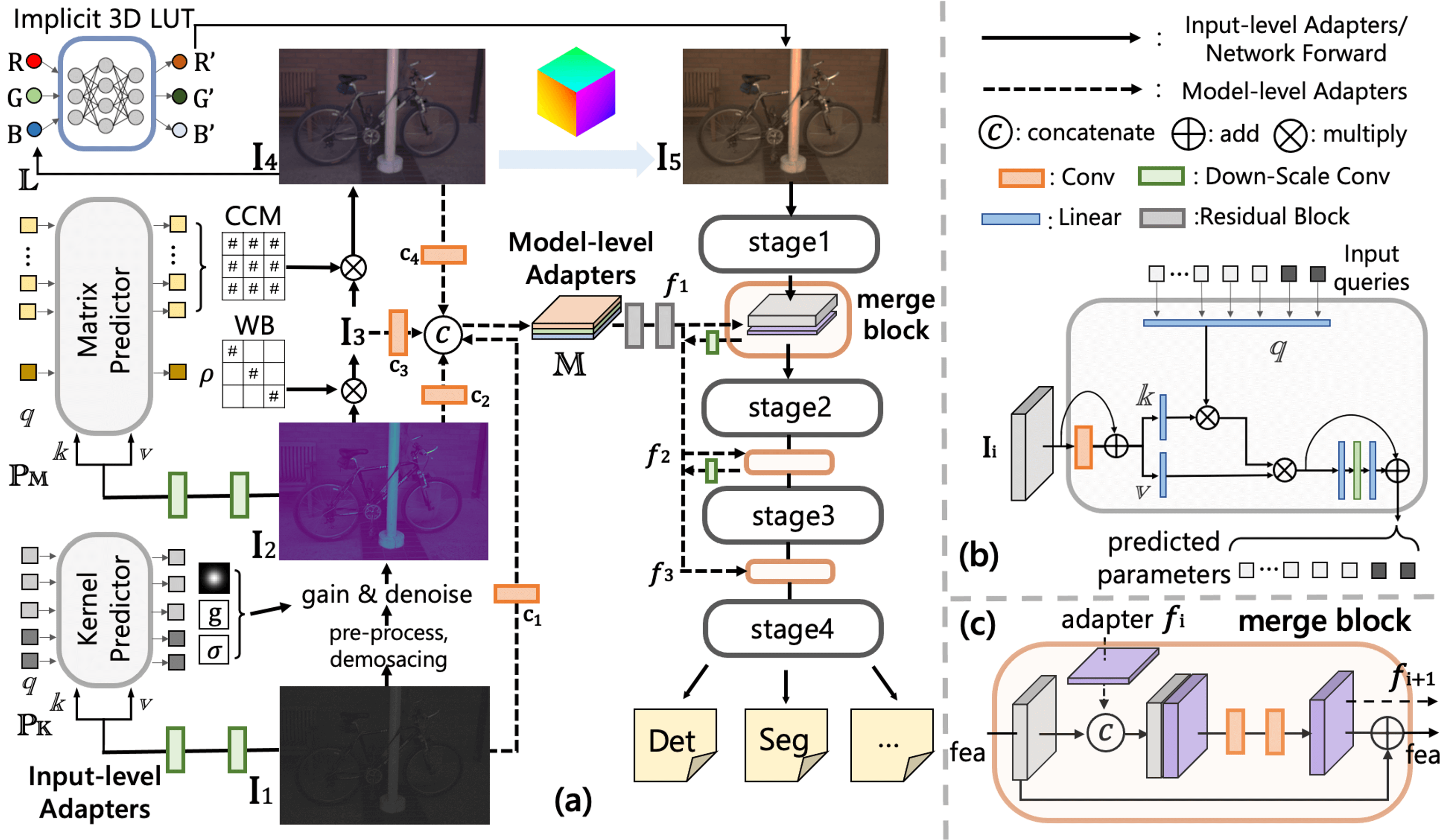}
    \caption{(a). Structure of RAW-Adapter. Solid line in left denotes input-level adapter's workflow and dotted line denotes model-level adapter's workflow, stage 1$\sim$4 means different stage of visual model backbone. (b). Detailed structure of kernel $\&$ matrix predictors $\mathbb{P_K}$, $\mathbb{P_M}$. (c). Detailed structure of model-level adapter $\mathbb{M}$'s merge block.}
    \label{fig:overview}
\end{figure*}

\subsection{Input-level Adapters}
\label{sec3.2input}

Fig.~\ref{fig:overview}(a) provides an overview of RAW-Adapter. The solid lines in Fig.~\ref{fig:overview}(a) left represent input-level adapters. Input-level adapters are designed to convert the RAW image $\mathbf{I}_1$ into the machine-vision oriented image $\mathbf{I}_5$. This process involves digital gain $\&$ denoise, demosacing, white balance adjustment, camera color matrix, and color manipulation.

We maintain the ISP design while simultaneously using query adaptive learning (QAL) to estimate key parameters in ISP stages. The QAL strategy is motivated by previous transformer models~\cite{DETR,cui2022you,cheng2021maskformer} and detailed structure is shown in Fig.~\ref{fig:overview}(b), input image $\mathbf{I}_{i\in(1,2)}$ would pass by 2 down-scale convolution blocks to generate feature, then feature pass by 2 linear layers to generate attention block's key $\mathbbm{k}$ and value $\mathbbm{v}$, while query $\mathbbm{q}$ is a set of learnable dynamic parameters, the ISP parameters would be predicted by self-attention calculation~\cite{vaswani2017attention}: 


\begin{equation}
    parameters = FFN(softmax(\frac{\mathbbm{q}  \cdot \mathbbm{k}^T}{\sqrt{d_k}}) \cdot \mathbbm{v}),
    \label{eq:predictor}
\end{equation}
where $FFN$ denotes the feed-forward network, includes 2 linear layers and 1 activation layer, the predicted parameters would keep the same length as query $\mathbbm{q}$. We defined 2 QAL blocks $\mathbb{P_K}$ and $\mathbb{P_M}$ to predict different part parameters.


The input RAW image $\mathbf{I}_1$ would first go through the pre-process  operations and a demosacing stage~\cite{Michael_eccv16,Mobile_Computational}, followed by subsequent ISP processes:

\subsubsection{Gain $\&$ Denoise:}


Denoising algorithms always take various factors into account, such as input photon number, ISO gain level, exposure settings.   
Here we first use QAL block $\mathbb{P_K}$ to predict a gain ratio $g$~\cite{brooks2019unprocessing,SID} to adapt $\mathbf{I}_1$ in  different lighting scenarios, followed by an adaptive anisotropic Gaussian kernel to suppress noise under various noise conditions, $\mathbb{P_K}$ will predict the appropriate Gaussian kernel $k$ for denoising to improve downstream visual tasks' performance, the predicted key parameters are the Gaussian kernel's major axis $r_1$, minor axis $r_2$ (see Fig.~\ref{fig:para} left), here we set the kernel angle $\theta$ to $0$ for simplification.  Gaussian kernel $k$ at pixel $(x, y)$ would be: 


\begin{equation}
    k(x,y) = exp(-(b_0  x^2 + 2b_1  x y + b_2y^2)), 
\end{equation}
where: 
\begin{equation}
    b_0 = \frac{cos(\theta)^2}{2{r_1}^2} + \frac{sin(\theta)^2}{2{r_2}^2}, b_1 = \frac{sin(2\theta)}{4{r_1}^2}((\frac{r_1}{r_2})^2 - 1), b_2 = \frac{sin(\theta)^2}{2{r_1}^2} + \frac{cos(\theta)^2}{2{r_2}^2}.  
\end{equation}

 After gain ratio $g$ and kernel $k$ process on image $\mathbf{I}_1$, $\mathbb{P_K}$ also predict a filter parameter $\sigma$ (initial at 0) to keep the sharpness and recover details of generated image $\mathbf{I}_2$. Eq.~\ref{eq:kernel_process} shows the translation from $\mathbf{I}_1$ to $\mathbf{I}_2$, where $\circledast$ denotes kernel convolution and filter parameter $\sigma$ is limited in a range of (0, 1) by a Sigmoid activation.  For more details please refer to our supplementary part Sec.~\ref{sec:detail_design}.

\begin{equation}
\begin{aligned}
    & \mathbb{P_K}(\mathbf{I}_{1}, \mathbbm{q}) \rightarrow k\left\{r_1, r_2, \theta \right\}, g, \sigma,  \\
    & \mathbf{I}_2' = (g \cdot \mathbf{I}_1) \circledast k,  \\
    & \mathbf{I}_2 = \mathbf{I}_2' + (g \cdot \mathbf{I}_1 - \mathbf{I}_2') \cdot \sigma .
\label{eq:kernel_process}
\end{aligned}
\end{equation}





\subsubsection{White Balance $\&$ CCM Matrix:}
\begin{figure*}[t]
    \centering
    \includegraphics[width=0.95\linewidth]{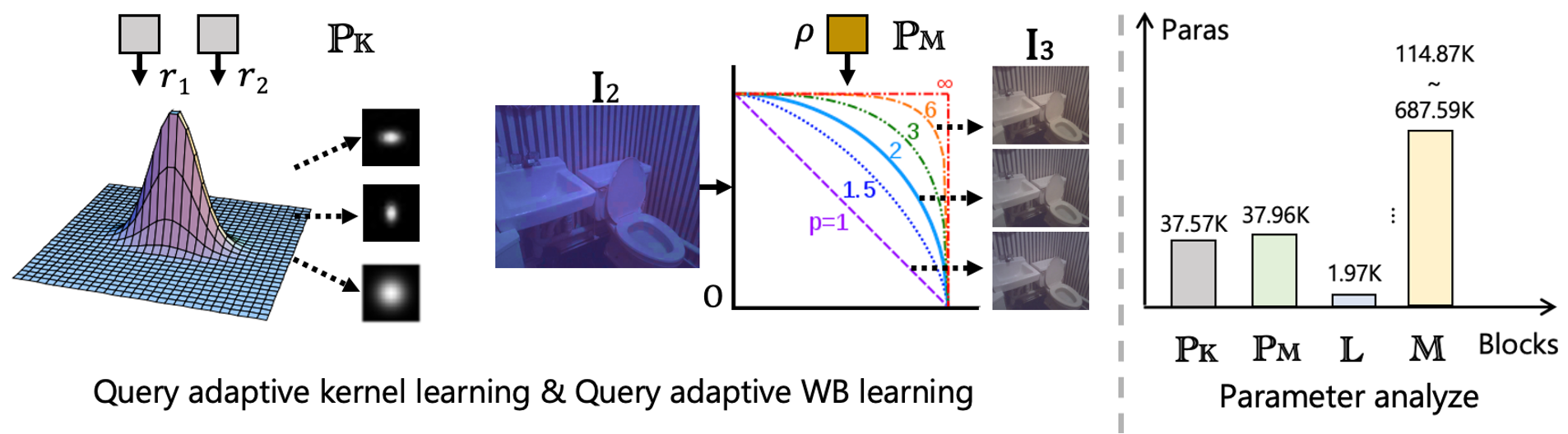}
    \caption{Left, we use query adaptive learning (QAL) to predict key parameters is ISP process. Right, we show RAW-Adapter different blocks' parameter.}
    \label{fig:para}
\end{figure*}

White balance (WB) mimics the color constancy of the human vision system (HVS) by aligning ``white'' color with the white object, resulting in the captured image reflecting the combination of light color and material reflectance~\cite{ICCV_MAET,Error_WB_Vision,brooks2019unprocessing}. In our work, we hope to find an adaptive white balance for different images under various lighting scenarios. Motivated by the design of Shades of Gray (SoG)~\cite{Shades_of_gray} WB algorithm, where gray-world WB and Max-RGB WB can be regarded as a subcase, we've employed a learnable parameter $\rho$ to replace the gray-world's L-1 average with an adaptive Minkowski distance average (see Eq.~\ref{eq:matrix_process}), the QAL block $\mathbb{P_M}$ predicts Minkowski distance's hyper-parameter $\rho$, a demonstration of various $\rho$  is shown in  Fig.~\ref{fig:para}. After finding the suitable $\rho$, $\mathbf{I}_2$ would multiply to white balance matrix to generate $\mathbf{I}_3$.



Color Conversion Matrix (CCM) within the ISP is constrained by the specific camera model. Here we standardize the CCM as a single learnable 3$\times$3 matrix, initialized as a unit diagonal matrix $\mathbf{E}_3$. The QAL block $\mathbb{P_M}$ predicts 9 parameters here, which are then added to $\mathbf{E}_3$ to form the final 3$\times$3 matrix $\mathbf{E}_{ccm}$. Then $\mathbf{I}_3$ would multiply to CCM $\mathbf{E}_{ccm}$ to generate $\mathbf{I}_4$, equation as follow:

\begin{equation}
\begin{aligned}
    &  \mathbb{P_M}(\mathbf{I}_{2}, \mathbbm{q}) \rightarrow \rho, \mathbf{E}_{ccm},\\
    & m_{i \in (r,g,b)} = \sqrt[\rho]{avg(\mathbf{I}_{2}(i)^\rho)} /  \sqrt[\rho]{avg((\mathbf{I}_{2})^\rho)}, \\
    & \mathbf{I}_{3} = \mathbf{I}_{2} * \begin{bmatrix}
   m_r &  &  \\
   & m_g &  \\
    &  & m_b
  \end{bmatrix}, \quad  \mathbf{I}_{4} = \mathbf{I}_{3} * \mathbf{E}_{ccm}.
\label{eq:matrix_process}
\end{aligned}
\end{equation}


\subsubsection{Color Manipulation Process:} In ISP, color manipulation process is commonly achieved through lookup table (LUT), such as 1D and 3D LUTs~\cite{Mobile_Computational}. Here, we consolidate color manipulation operations into a single 3D LUT, adjusting the color of image $\mathbf{I}_4$ to produce output image $\mathbf{I}_5$. Leveraging advancements in LUT techniques, we choose the latest neural implicit 3D LUT (NILUT)~\cite{conde2024nilut}, for its speed efficiency and ability to facilitate end-to-end learning in our pipeline~\footnote{To save memory occupy, we change the channel number 128 in~\cite{conde2024nilut} to 32.}. We denote NILUT~\cite{conde2024nilut} as $\mathbb{L}$, $\mathbb{L}$ maps the input pixel intensities $R, G, B$ to a continuous coordinate space, followed by the utilization of implicit neural representation (INR)~\cite{sitzmann2019siren}, which involves using a multi-layer perceptron (MLP) network to map the output pixel intensities to $R', G', B'$:

\begin{equation}
    \mathbf{I}_5(R', G', B') =  \mathbb{L}(\mathbf{I}_4(R, G, B)).
\end{equation}

Image $\mathbf{I}_5$ obtained through image-level adapters will be forwarded to the downstream network's backbone. Furthermore, $\mathbf{I}_5$'s features obtained in backbone will be fused with model-level adapters, which we will discuss in detail.


\subsection{Model-level Adapters}
\label{sec3.3model}

Input-level adapters guarantee the production of machine vision-oriented image $\mathbf{I}_5$ for high-level models. However, information at the ISP stages ($\mathbf{I}_1 \sim \mathbf{I}_4$) is almost overlooked. Inspired by adapter design in current NLP and CV area~\cite{Prompt_NLP,Vit_adapter,VPT_ECCV2022}, we employ the prior information from the ISP stage as model-level adapters to guide subsequent models' perception. Additionally, model-level adapters promote a tighter integration between downstream tasks and the ISP stage. Dotted lines in Fig.~\ref{fig:overview}(a) represent model-level adapters.

Model-level adapters $\mathbb{M}$ integrate the information from ISP stages to the network backbone. As shown in Fig.~\ref{fig:overview}(a), we denote the stage $1 \sim 4$ as different stages in network backbone, image $\mathbf{I}_5$ would pass by backbone stages $1 \sim 4$ and then followed by detection or segmentation head. We utilize convolution layer $c_i$ to extract features from $\mathbf{I}_1$ through $\mathbf{I}_4$, these extracted features are concatenated as $\mathbb{C}(\mathbf{I}_{1\sim4}) = \mathbb{C}(c_1(\mathbf{I}_1), c_2(\mathbf{I}_2), c_3(\mathbf{I}_3), c_4(\mathbf{I}_4))$. Subsequently, $\mathbb{C}(\mathbf{I}_{1\sim4})$ go through two residual blocks~\cite{he2016resnet} and generate adapter $f_1$, $f_1$ would merge with stage $1$'s feature by a merge block, detail structure of merge block is shown in Fig.~\ref{fig:overview}(c), which roughly includes concatenate process and a residual connect, finally merge block would output network feature for stage $2$ and an additional adapter $f_2$. Then process would repeat in stage $2$ and stage $3$ of network backbone. We collectively refer to all structures related to model-level adapters as $\mathbb{M}$.


We show RAW-Adapter's different part parameter number in Fig.~\ref{fig:para} right, including input-level adapters $\{ \mathbb{P_K}$ (37.57K), $\mathbb{P_M}$ (37.96K), $\mathbb{L}$ (1.97K)$\}$ and model level adapters $\mathbb{M}$ (114.87K $\sim$ 687.59K), model level adapters' parameter number depend on following network. Total parameter number of RAW-Adapter is around 0.2M to 0.8M, much smaller than following backbones ($\sim$25.6M in ResNet-50~\cite{he2016resnet}, $\sim$197M in Swin-L~\cite{liu2021Swin}), also smaller than previous SOTA methods like SID~\cite{SID} (11.99M) and Dirty-Pixel~\cite{steven:dirtypixels2021} (4.28M). Additionally, subsequent experiments will demonstrate that the performance of RAW-Adapter on backbone networks with fewer parameters is even superior to that of previous algorithms on backbone networks with higher parameters.

%
%

\section{Experiments}

\subsection{Dataset and Experimental Setting}

\subsubsection{Dataset.} We conducted experiments on object detection and semantic segmentation tasks, utilizing a combination of various synthetic and real-world RAW image datasets, an overview of the datasets can be found in Table~\ref{tab:dataset}. 

For the object detection task, we take 2 open source real-world dataset PASCAL RAW~\cite{omid2014pascalraw} and LOD~\cite{LOD_BMVC2021}. PASCAL RAW~\cite{omid2014pascalraw} is a normal-light condition dataset with 4259 RAW images, taken by a Nikon D3200 DSLR camera with 3 object classes. To verify the generalization capability of RAW-Adapter across various lighting conditions, we additionally synthesized low-light and overexposure datasets based on the PASCAL RAW dataset, named PASCAL RAW (dark) and PASCAL RAW (over-exp) respectively, the synthesized datasets are identical to the PASCAL RAW dataset in all aspects except for brightness levels. For PASCAL RAW (dark) and PASCAL RAW (over-exp) synthesis, the light intensity and environment irradiance exhibit a linear relationship on RAW images, thus we employed the synthesis method from the previous work~\cite{ICCV_MAET,ISP-Teacher}:

\begin{table}[t]
\caption{Dataset and framework setting in our experiments.}
\label{tab:dataset}
\large
\renewcommand\arraystretch{1.3}
\begin{adjustbox}{max width = \linewidth}
\begin{tabular}{c|ccc|cc}
\hline
\hline
       & \multicolumn{1}{c|}{PASCAL RAW} & \multicolumn{1}{c|}{\begin{tabular}[c]{@{}c@{}}PASCAL RAW \\ (dark/over-exp)\end{tabular}}  & LOD                  & \multicolumn{1}{c|}{ADE20K RAW} & \multicolumn{1}{c}{\begin{tabular}[c]{@{}c@{}}ADE20K RAW \\ (dark/over-exp)\end{tabular}} \\ \hline
Task   & \multicolumn{3}{c|}{Object Detection}                                                                                                                                                                                              & \multicolumn{2}{c}{Semantic Segmentation}                                                                                                                                              \\ \hline
Number & \multicolumn{2}{c|}{4259}                                                                                                                                                                                   & 2230                 & \multicolumn{2}{c}{27,574}                                                                                                                                                             \\ \hline
Type   & \multicolumn{1}{c|}{real-world} & \multicolumn{1}{c|}{synthesis}                                                                                                                                            & real-world           & \multicolumn{2}{c}{synthesis}                                                                                                                                                          \\ \hline
Sensor & \multicolumn{2}{c|}{Nikon D3200 DSLR}                                                                                                                                                                       & Canon EOS 5D Mark IV & \multicolumn{2}{c}{-}                                                                                                                                              \\ \hline
Framework & \multicolumn{3}{c|}{RetinaNet~\cite{lin2017focal_loss} $\&$ Sparse-RCNN~\cite{sparse_rcnn}}                                                                                                                                                                     & \multicolumn{2}{c}{Segformer~\cite{xie2021segformer}} 

\\ \hline
Backbone & \multicolumn{3}{c|}{ResNet~\cite{he2016resnet}}                                                                                                                                                                     & \multicolumn{2}{c}{MIT~\cite{xie2021segformer}} 

\\
\hline
pre-train & \multicolumn{5}{c}{ImageNet~\cite{COCO_dataset} pre-train weights}                                                                                                                                                                
\\ \hline
\hline
\end{tabular}
\end{adjustbox}

\end{table}

\begin{equation}
\begin{aligned}
    &x_{n} \sim N(\mu = lx, \sigma^{2} = \delta_{r}^2 + \delta_{s}lx)\\
    &y = lx + x_{n},
\end{aligned}
\end{equation}
where $x$ denotes the original normal-light RAW image and $y$ denotes degraded RAW image, $\delta_{s} = \sqrt{S}$ denotes shot noise while $\sqrt{S}$ is signal of the sensor, $\delta_{r}$ denotes read noise, $l$ denotes the 
light intensity parameter which randomly chosen from [0.05, 0.4] in PASCAL RAW (dark), and randomly chosen from [2.5, 3.5] in PASCAL RAW (over-exp). Additionally, we follow PASCAL RAW~\cite{omid2014pascalraw}'s dataset split to separate the training set and test set.

Meanwhile, LOD~\cite{LOD_BMVC2021} is a real-world dataset with 2230 low-light condition RAW images, taken by a Canon EOS 5D Mark IV camera with 8 object classes, we take 1800 images as training set and the other 430 images as test set.


For the semantic segmentation task, we utilized the widely-used sRGB dataset ADE20K~\cite{ADE20K} to generate RAW dataset. Leveraging state-of-the-art unprocessing methods InvISP~\cite{invertible_ISP}, we synthesized RAW images corresponding to the ADE20K sRGB dataset. Using InvISP, we projected input sRGB images into RAW format, effectively creating an ADE20K RAW dataset. To simulate various lighting conditions, we employed the same synthesis method from PASCAL RAW (dark/over-exp) to generate low-light and over-exposure RAW images, name as ADE20K RAW (dark/over-exp), training $\&$ test split is same as ADE20K.

\subsubsection{Implement Details.} We build our framework based on the open-source computer vision toolbox~\texttt{mmdetection}~\cite{chen2019mmdetection} and~\texttt{mmsegmentation}~\cite{mmseg2020}, both object detection tasks and semantic segmentation tasks are initialed with ImageNet pre-train weights (see Table.~\ref{tab:dataset}), and we apply the data augmentation pipeline in the default setting, mainly include random crop, random flip, and multi-scale test, etc. For the object detection task, we adopt the 2  mainstream object detectors: RetinaNet~\cite{lin2017focal_loss} and Sparse-RCNN~\cite{sparse_rcnn} with ResNet~\cite{he2016resnet} backbone. For the semantic segmentation task, we choose to use the mainstream segmentation framework Segformer~\cite{xie2021segformer} with MIT~\cite{xie2021segformer} backbone.

\subsubsection{Comparison Methods.}
We conducted comparative experiments with the current state-of-the-art (SOTA) algorithms, including various open-sourced ISP methods~\cite{Michael_eccv16,SID,invertible_ISP,RAW-to-sRGB_ICCV2021,jincvpr23dnf} and joint-training method DirtyPixels~\cite{steven:dirtypixels2021}, among these Karaimer~\textit{et al.}~\cite{Michael_eccv16} is a traditional ISP method with various human manipulate steps, and InvISP~\cite{invertible_ISP} $\&$ Lite-ISP~\cite{RAW-to-sRGB_ICCV2021} $\&$ SID~\cite{SID} $\&$ DNF~\cite{jincvpr23dnf} are current SOTA network-based ISP methods, where SID~\cite{SID} and DNF~\cite{jincvpr23dnf} is especially for the low-light condition RAW data. For fairness compared with the above ISP methods, both the training and test RAW data are rendered using the respective compared ISP. DirtyPixels~\cite{steven:dirtypixels2021} is the current SOTA joint-training method which uses a stack of residual UNet~\cite{UNet} as a low-level processor and joint optimization with the following task-specific networks. 
For fairness, all comparison methods adopt the same data augmentation process and the same training setting, we will introduce detailed experiment results in the following section.


\begin{figure*}[t]
    \centering
    \includegraphics[width=1.00\linewidth]{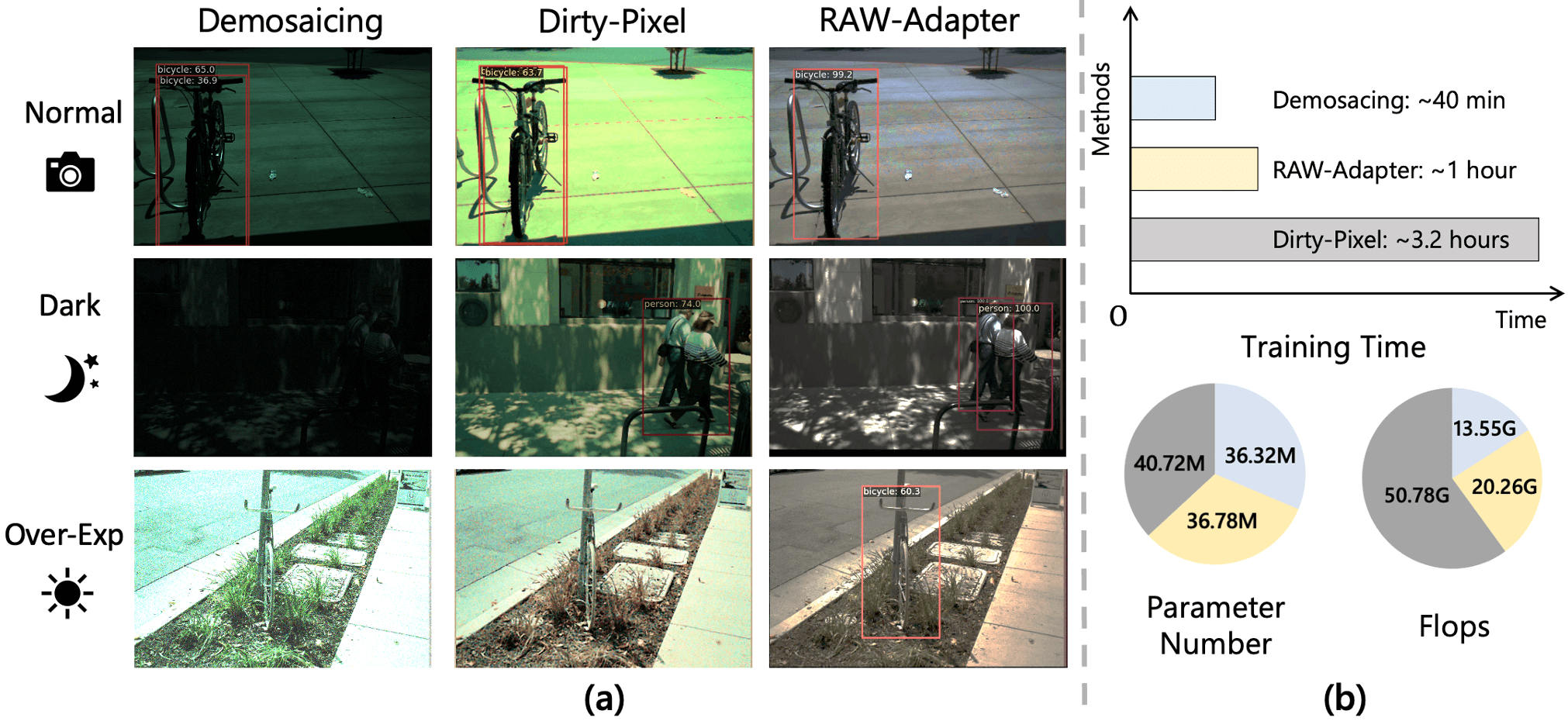}
    \caption{(a). Detection performance on \textbf{PASCAL RAW}~\cite{omid2014pascalraw} (normal/dark/over-exp). (b). Efficiency comparison (blue: vanilla, yellow: RAW-Adapter, gray: Dirty-Pixel~\cite{steven:dirtypixels2021}).}
    \label{fig:detection}
\end{figure*}

\begin{table*}[t]
  \centering
  \caption{Comparison results on \textbf{PASCAL RAW} dataset~\cite{omid2014pascalraw}, we take~\cite{lin2017focal_loss} with (a). ResNet-18 and (b). ResNet-50 backbone, \textbf{bold} denotes the best result.}
  \label{tab:Pascal_RAW}
  
  \begin{subtable}[t]{0.48\linewidth}
  \centering
  
  \caption{ResNet-18.}
  \label{tab:ResNet-18}
  
  \resizebox{\linewidth}{!}{%
  \begin{tabular}{c | c c c c}
    \toprule
    \diagbox{Method}{mAP}  & normal  & over-exp  & \ dark \  \\
    \midrule
    Default ISP & 88.3 & - & -  \\
    
    Demosacing & 87.7 & 87.7 & 80.3  \\ 
    \midrule
     Karaimer~\textit{et al.}~\cite{Michael_eccv16} & 88.1 & 85.6 & 78.8\\ 
     Lite-ISP~\cite{RAW-to-sRGB_ICCV2021} & 85.2 & 84.2 & 71.9 \\ 
     InvISP~\cite{invertible_ISP}  & 85.4 & 86.6 & 70.9 \\
    \midrule
    SID~\cite{SID} & - & - & 78.2\\ 
     DNF~\cite{jincvpr23dnf} & - & - & 81.1 \\ 
    \midrule
    Dirty-Pixel   & 88.6 & 88.0 & 80.8 \\ 
    \textbf{RAW-Adapter} & \textbf{88.7} & \textbf{88.7} & \textbf{82.5} \\ 
     
    \bottomrule
    \end{tabular}}
  \end{subtable} \hfill
  \begin{subtable}[t]{0.488\linewidth}
  \centering
  \caption{ResNet-50.}
  \label{tab:blur}
  
  \resizebox{\linewidth}{!}{%
   \begin{tabular}{c | c c c c}
    \toprule
    \diagbox{Method}{mAP}  & normal  & over-exp  & \ dark \  \\

    \midrule
    Default ISP & 89.6
    & - & -  \\
    Demosacing & 89.2 & 88.8 & 82.6  \\ 
    \midrule
     Karaimer~\textit{et al.}~\cite{Michael_eccv16} & 89.4 & 86.8 &  79.6\\ 
     Lite-ISP~\cite{RAW-to-sRGB_ICCV2021} & 88.5 & 85.1 & 73.5 \\ 
     InvISP~\cite{invertible_ISP}  & 87.6 & 87.3 & 74.7 \\
    \midrule
    SID~\cite{SID} & - & - & 81.5\\ 
     DNF~\cite{jincvpr23dnf} & - & - & 82.8 \\ 
    \midrule
    Dirty-Pixel   & \textbf{89.7} & 89.0 & 83.6 \\ 
    \textbf{RAW-Adapter} & \textbf{89.7} & \textbf{89.5} & \textbf{86.6} \\ 
     
    \bottomrule
    \end{tabular}
  }
  \end{subtable}
\end{table*}

\begin{table}[t]
\caption{Comparison with ISP methods and Dirty-Pixel on \textbf{LOD} dataset~\cite{LOD_BMVC2021}. We show the detection performance (mAP) $\uparrow$ of RetinaNet (R-Net)~\cite{lin2017focal_loss} and Sparse-RCNN (Sp-RCNN)~\cite{sparse_rcnn}, \textbf{bold} denotes the best
result.}
\label{tab:LOD}
\renewcommand\arraystretch{1.3}
\resizebox{\linewidth}{!}{%
\begin{tabular}{c|cccc}
\toprule
Methods & Demosaicing  & Default ISP
  & Karaimer~\textit{et al.}~\cite{Michael_eccv16} 
& InvISP~\cite{invertible_ISP}      \\ \hline
mAP(R-Net)      & 58.5 &  58.4  & 54.4   &    56.9                     \\ \hline
 mAP(Sp-RCNN)     & 57.7 & 53.9  &    52.2   &  49.4                    \\ \midrule
Methods & SID~\cite{SID}  & Dirty-Pixel~\cite{steven:dirtypixels2021}   &     \textbf{RAW-Adapter} (w/o $\mathbb{M}$)      & \textbf{RAW-Adapter}     \\ \hline
mAP(R-Net)      & 49.1 & 61.6   & 61.6        & \textbf{62.1}                       \\ \hline
 mAP(Sp-RCNN)      & 43.1  & 58.8         & 58.6         & \textbf{59.2}               \\ \bottomrule
\end{tabular}}
\end{table}

\subsection{Object Detection Evaluation}

For object detection task on PASCAL RAW~\cite{omid2014pascalraw} dataset, we adopt RetinaNet~\cite{lin2017focal_loss} with different size ResNet~\cite{he2016resnet} backbones (ResNet-18, ResNet-50), all the models are trained on a single NVIDIA Tesla V100 GPU with SGD optimizer, the batch size is set to 4, training images are cropped
into range of (400, 667) and training epochs are set to 50. Table.~\ref{tab:Pascal_RAW} shows the detection results with (a). ResNet-18 and (b). ResNet-50 backbone, with comparisons of demosaiced RAW data (``Demosacing''), camera default ISP in~\cite{omid2014pascalraw}, various current SOTA ISP solutions~\cite{Michael_eccv16,RAW-to-sRGB_ICCV2021,SID,jincvpr23dnf} and Dirty-Pixel~\cite{steven:dirtypixels2021}, we can observe that sometimes ISP algorithms in normal and over-exposure conditions may even degraded the detection performance, and ISP in dark scenarios sometimes could improves detection performance, additionally manual design ISP solution~\cite{Michael_eccv16} even out-perform the deep learning solutions. Previous joint-training method Dirty-Pixel~\cite{steven:dirtypixels2021} can improve detection performance in all scenarios. Overall, our RAW-Adapter method achieves the best performance and even outperforms some ISP algorithms utilizing ResNet-50 backbone when employing ResNet-18 backbone, which can significantly reduce computational load and model parameters.

Visualization of PASCAL RAW's detection results are shown in Fig.~\ref{fig:detection}(a), we show the detection results under different lightness conditions, background are 
images generated from Dirty-Pixel~\cite{steven:dirtypixels2021} and RAW-Adapter ($\mathbf{I}_5$). Our method could achieve satisfactory detection results across different lightness, while other methods face false alarm and miss detect. Efficiency comparison with Dirty-Pixel is shown in Fig.~\ref{fig:detection}(b), our algorithm has achieved significant improvements in both training time acceleration and savings in Flops/Parameter Number. 

Detection performance on LOD~\cite{LOD_BMVC2021} is shown in Table.~\ref{tab:LOD}, we adopt two detector RetinaNet~\cite{lin2017focal_loss} and Sparse-RCNN~\cite{sparse_rcnn}, both with ResNet-50~\cite{he2016resnet} backbone, training epochs are set to 35, Sparse-RCNN~\cite{sparse_rcnn} are trained by 4 GPUs with Adam optimizer. With comparing of various ISP methods~\cite{Michael_eccv16,RAW-to-sRGB_ICCV2021,SID,jincvpr23dnf}, Dirty-Pixel~\cite{steven:dirtypixels2021} and RAW-Adapter with only input-level adapters (w/o $\mathbb{M}$). Table.~\ref{tab:LOD} show that our algorithm can achieve the best performance on both two detectors.

\subsection{Semantic Segmentation Evaluation}

For semantic segmentation on ADE20K RAW dataset, we choose Segformer~\cite{xie2021segformer} as the segmentation framework with the different size MIT~\cite{xie2021segformer} backbones (MIT-B5, MIT-B3, MIT-B0), all the models are trained on 4 NVIDIA Tesla V100 GPU with Adam optimizer, the batch size is set to 4, training images are cropped into $512\times512$ and training iters are set to $80000$. We make comparison with single demosacing, various ISP methods~\cite{Michael_eccv16,invertible_ISP,RAW-to-sRGB_ICCV2021,jincvpr23dnf,SID} and Dirty-Pixel~\cite{steven:dirtypixels2021}, and RAW-Adapter with only input-level adapters (w/o $\mathbb{M}$).

\begin{figure*}[t]
    \centering
    \includegraphics[width=0.95\linewidth]{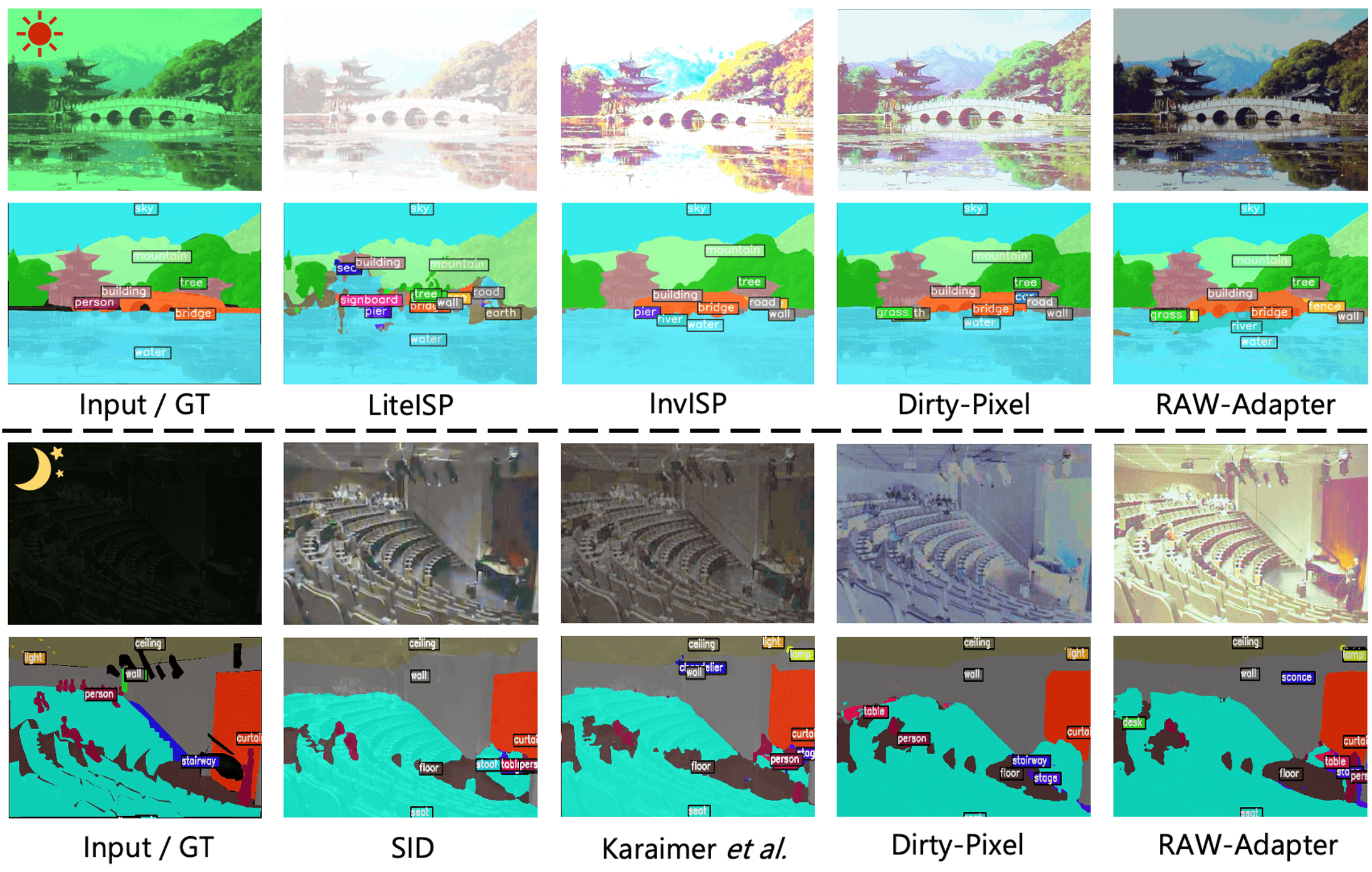}
    \caption{Semantic segmentation results on over-exposure and low-light RAW data, compare with various ISP methods~\cite{RAW-to-sRGB_ICCV2021,invertible_ISP,SID,Michael_eccv16} and Dirty-Pixel~\cite{steven:dirtypixels2021}.}
    \label{fig:segmentation}
   
\end{figure*}

Comparison results are shown in Table.~\ref{tab:segmentation}, where we compare both efficiency (parameters$\downarrow$, inference time$\downarrow$) and performance (mIOU$\uparrow$), inference time is calculated on a single Tesla V100 GPU. We first show the results on vanilla demosaiced RAW data, then we show the results on RAW data processed by various ISP methods~\cite{Michael_eccv16,RAW-to-sRGB_ICCV2021,invertible_ISP,jincvpr23dnf,SID}, followed by results of Dirty-Pixel~\cite{steven:dirtypixels2021}, and our method without and with model-level adapter $\mathbb{M}$. For Table.~\ref{tab:segmentation}, we can find that directly employing ISP methods sometimes fails to enhance the performance of downstream segmentation tasks, particularly in low-light conditions. Joint-training method Dirty-Pixel~\cite{steven:dirtypixels2021} could improve down-stream performance and sometimes even degrade segmentation performance, but increase more network parameters. Overall, our proposed RAW-Adapter could effectively enhance performance while adding a slight parameter count and improving inference efficiency. RAW-Adapter on backbones with fewer parameters can even outperform previous methods on backbones with more parameters (\textit{i.e.} The RAW-Adapter, employing the MIT-B3 backbone, achieves a dark condition mIOU of 37.62, surpassing other methods utilizing the MIT-B5 backbone despite having fewer than $\sim$ 36M parameters.).

The visualization of segmentation results are shown in Fig.~\ref{fig:segmentation}, where row 1 and row 3 depict images rendered by different ISPs~\cite{RAW-to-sRGB_ICCV2021,invertible_ISP,SID,Michael_eccv16} and images generated from Dirty-Pixel~\cite{steven:dirtypixels2021} and RAW-Adapter ($\mathbf{I}_5$), row 2 and row 4 denote our segmentation results appear superior compared to other methods.
We were even pleased to discover that, although we do not add any human-vision-oriented loss function to constrain the model, RAW-Adapter still produces visually satisfactory image $\mathbf{I}_5$ (see Fig.~\ref{fig:segmentation}). This may be attributed to our preservation of traditional ISP stages in input-level adapters. Due to page limitations, we refer to more experimental results, visualization results, and ablation analysis of our method in the supplementary part.

\begin{table}[t]
\caption{Comparison with previous methods on \textbf{ADE 20K RAW} (normal/over-exp/dark). \textbf{Bold} denotes the best result while \underline{underline} denotes second best result.}
\label{tab:segmentation}
\renewcommand\arraystretch{1.2}
\resizebox{\linewidth}{!}{%
\begin{tabular}{c|c|c|c|c|c|c}
\toprule
\toprule
  & backbone   &  params(M) $\downarrow$ & \begin{tabular}[c]{@{}c@{}}inference\\  time(s) $\downarrow$ \end{tabular} & \begin{tabular}[c]{@{}c@{}}mIOU $\uparrow$ \\ (normal)\end{tabular} & \begin{tabular}[c]{@{}c@{}}mIOU $\uparrow$\\ (over-exp)\end{tabular} & \  \begin{tabular}[c] {@{}c@{}}mIOU $\uparrow$\\ (dark)\end{tabular} \  \\ \hline
Demosacing  & \multirow{7}{*}{MIT-B5} &      \multirow{6}{*}{82.01}                                                     &    0.105   &  47.47   &  45.69    &    37.55        \\ \cline{1-1} \cline{4-7} 
Karaimer \textit{et al.}~\cite{Michael_eccv16} &                         &          &     0.525          & 45.48       & 42.85                                                    & 37.32                                               \\ \cline{1-1} \cline{4-7} 
InvISP~\cite{invertible_ISP}                                                                  &         &                & 0.203                                            & 47.82                                                   & 44.30                                                     & 4.03                                                \\ \cline{1-1} \cline{4-7} 
LiteISP~\cite{RAW-to-sRGB_ICCV2021}                                                                 &                         &                                                             &     0.261       & 43.22                                                   & 42.01                                                     & 5.52                                                 \\ \cline{1-1} \cline{4-7} 
DNF~\cite{jincvpr23dnf}                                                                     &                         &         &   0.186                                          & -                                                       & -                                                         & 35.88                                              \\ \cline{1-1} \cline{4-7} 
SID~\cite{SID}                                                                     &                         &      &    0.312                                       & -                                                       & -                                                         & 37.06 \\ \cline{1-1} \cline{3-7} 
\multirow{3}{*}{Dirty-Pixel~\cite{steven:dirtypixels2021}}     &   &  86.29  &   0.159  & \underline{47.86}  & \underline{46.50}  & 38.02 \\ \cline{2-7}  & \cellcolor[HTML]{ECF4FF}MIT-B3  & \cellcolor[HTML]{ECF4FF}48.92  & \cellcolor[HTML]{ECF4FF}0.098  & \cellcolor[HTML]{ECF4FF}46.19 & \cellcolor[HTML]{ECF4FF}44.13  & \cellcolor[HTML]{ECF4FF}36.93 \\ \cline{2-7} 
& \cellcolor{gray!10}MIT-B0 & \cellcolor{gray!10}8.00 & \cellcolor{gray!10}\underline{0.049}  & \cellcolor{gray!10}34.43 &  \cellcolor{gray!10}31.10  &      \cellcolor{gray!10}24.89                                                \\ \hline
\multirow{3}{*}{\begin{tabular}[c]{@{}c@{}}\textbf{RAW-Adapter} \\ (w/o $\mathbb{M}$)\end{tabular}}  & MIT-B5    &  82.09   &   0.148  &  47.83  &  46.48 & \underline{38.41} \\ \cline{2-7} & \cellcolor[HTML]{ECF4FF}MIT-B3   &   \cellcolor[HTML]{ECF4FF}44.72  & \cellcolor[HTML]{ECF4FF}0.086  & \cellcolor[HTML]{ECF4FF}46.22  & \cellcolor[HTML]{ECF4FF}44.00  &  \cellcolor[HTML]{ECF4FF}37.60  \\ \cline{2-7}    & \cellcolor{gray!10}MIT-B0  &  \cellcolor{gray!10}\textbf{3.80}  &  \cellcolor{gray!10}\textbf{0.032} &  \cellcolor{gray!10}34.66 &   \cellcolor{gray!10}31.82   & \cellcolor{gray!10}23.99 \\ \hline

\multirow{3}{*}{\textbf{RAW-Adapter}}  & MIT-B5    &  82.31   &  0.167  &  \textbf{47.95}  &  \textbf{46.62}  &  \textbf{38.75} \\ \cline{2-7} & \cellcolor[HTML]{ECF4FF}MIT-B3   &  \cellcolor[HTML]{ECF4FF}45.16   & \cellcolor[HTML]{ECF4FF}0.102  & \cellcolor[HTML]{ECF4FF}46.57  & \cellcolor[HTML]{ECF4FF}44.19  & \cellcolor[HTML]{ECF4FF}37.62   \\ \cline{2-7}    & \cellcolor{gray!10}MIT-B0       &  \cellcolor{gray!10}\underline{3.87}   & \cellcolor{gray!10}0.053  & \cellcolor{gray!10}34.72 & \cellcolor{gray!10}31.91  & \cellcolor{gray!10}25.06 \\

\bottomrule
                                                                        \bottomrule
\end{tabular}}

\end{table}

\section{Conclusion}

In this paper, we introduce RAW-Adapter, an effective solution for adapting pre-trained sRGB models to camera RAW data. 
With input-level adapters and model-level adapters working in tandem, RAW-Adapter effectively forwards RAW images from various lighting conditions to downstream vision tasks, our method has achieved state-of-the-art (SOTA) results across multiple datasets.

For future research directions, we believe it is feasible to design a unified model capable of adapting to RAW visual tasks under different lighting conditions, without the need for retraining in each lighting scenario like RAW-Adapter. Additionally, designing multi-task decoders can enable the accommodation of various tasks, facilitating more effective integration of different visual tasks on RAW images with large-scale models.

\clearpage

\section*{Acknowledgements}
This research is partially supported by JST Moonshot R$\&$D Grant Number JPMJPS2011, CREST Grant Number JPMJCR2015 and Basic Research Grant (Super AI) of Institute for AI and Beyond of the University of Tokyo.

%
%
\bibliographystyle{splncs04}
\bibliography{main}

\newpage
\appendix

\renewcommand\thesection{\Alph{section}}
\renewcommand\thetable{\Alph{section}\arabic{table}} 
\renewcommand\thefigure{\Alph{section}\arabic{figure}}

\section{ISP revisited}
\label{sec3.1ISP_revisited}
Here we give a brief review of the digital camera image formation process, from camera sensor RAW data to output sRGB, please refer to Fig.~\ref{fig:motivation}(a) for an illustrative diagram, the ISP steps mainly include:

\textit{(a). Pre-processing} involves some pre-process operations such as BlackLevel adjustment, WhiteLevel adjustment, and lens shading correction.

\textit{(b). Noise reduction} eliminates noise and keeps the visual quality of image, this step is closely related to exposure time and camera ISO settings~\cite{Wei_2020_CVPR,Dancing_under_light}.

\textit{(c). Demosaicing} is used to reconstruct a 3-channel color image from a single-channel RAW, executed through interpolation of the absent values in the Bayer pattern, relying on neighboring values in the CFA.

\textit{(d). White Balance}  simulates the color constancy of human visual system (HVS). An auto white balance (AWB) algorithm estimates the sensor's response to illumination of the scene and corrects RAW data.

\textit{(e). Color Space Transformation} mainly includes two steps, first is mapping white balanced pixel to un-render color space (\textit{i.e.} CIEXYZ), and the second is mapping un-render color space to the display-referred color space (\textit{i.e.} sRGB), typically each use a 3$\times$3 matrix based on specific camera~\cite{Mobile_Computational}.


\textit{(f). Color and Tone Correction} are often implemented using 3D and 1D lookup tables (LUTs), while tone mapping also compresses pixel values.

\textit{(g). Sharpening}  enhances image details by unsharp masking or deconvolution.

We refer other detailed steps such as digital zoom and gamma correction to previous works~\cite{Michael_eccv16,Mobile_Computational,ISP_2005}. 
Meanwhile, in the ISP pipeline, many other operations prioritize the quality of the generated image rather than its performance in machine vision tasks.
Therefore, for specific adapter designs, we selectively omit certain steps and focus on including the steps mentioned above. We provide detailed explanations in the Sec.~\ref{sec3.2input} and Sec.~\ref{sec:detail_design}.

\section{Impact of Different Blocks}

We conducted ablation experiments to assess the effectiveness of different stages in RAW-Adapter. The experiments were designed on the PASCAL dataset with RetinaNet~\cite{lin2017focal_loss} (ResNet-50 backbone), covering normal, dark, and over-exposed conditions. The results are presented in Table.~\ref{tab:ablation_block}, we can find that the kernel predictor $\mathbb{P_K}$ exhibits significant improvements in dark scenarios (+2.4), attributable to the gain ratio $g$ and denoising processes, but it doesn't seem to be of much help in both overexposed and normal scenes (+0.0), this might be due to the current kernel-based denoising methods being too simplistic and eliminating some detail information. Meanwhile the implicit LUT $\mathbb{L}$ does not show improvement under over-exposed and low-light conditions  but proves effective in normal light condition. Finally, the model-level adapters $\mathbb{M}$ and matrix predictor $\mathbb{P_M}$ yield performance improvements across all scenarios.

\begin{table}[t]
\caption{Ablation analyze on RAW-Adapter's model structure.}
\label{tab:ablation_block}
\centering
\renewcommand\arraystretch{1.3}
\resizebox{10.5cm}{!}{
\begin{tabular}{c|ccccc|c|c|c}
\toprule
\toprule
\multirow{6}{*}{blocks} & \ base \ & \ $\mathbb{P_K}$ \ & \ $\mathbb{P_M}$ \ & \ $\mathbb{L}$  \ & \ $\mathbb{M}$ \ & \ mAP (normal) \ & \ mAP (over-exp) \ & \ \ mAP (dark) \ \ \\ \cline{2-9} 
                        & \checkmark     &        &        &     &     &   
                        89.2  &    88.8     &  82.6 \\
                        & \checkmark     & \checkmark   &        &     &     &  
                        89.2 (+0.0)    &   88.8 (+0.0)  &  85.0 (+2.4) \\
                        & \checkmark     & \checkmark   & \checkmark   &     &     & 89.5 (+0.2) &   89.0 (+0.2)   &  86.2 (+3.6)    \\
                        & \checkmark     & \checkmark   & \checkmark   & \checkmark&     &   
                        89.4 (- 0.1)  &   89.0 (+0.2)    &  86.3 (+3.7) \\
                        & \checkmark     & \checkmark   & \checkmark   & \checkmark& \checkmark&   
                        89.7 (+0.5)  & 89.5 (+0.7)    &   86.6 (+4.0) \\ \bottomrule \bottomrule
\end{tabular}}
\end{table}

\section{Detailed Design of Input-level Adapters}
\label{sec:detail_design}

In the main text of our paper, we outlined that the input level adapters of RAW-Adapter comprise three components: the kernel predictor $\mathbb{P_K}$, the matrix predictor $\mathbb{P_M}$, and the neural implicit 3D LUT $\mathbb{L}$. In this section, we will provide a detailed explanation of how to set the parameter ranges for input-level adapters, along with conducting some results analysis. 

The kernel predictor $\mathbb{P_K}$ is responsible for predicting five ISP-related parameters, including the \ding{172} gain ratio $g$, the  Gaussian kernel \ding{173} $k$'s major axis radius $r_1$, \ding{174} $k$'s minor axis radius $r_2$, and the \ding{175} sharpness filter parameter $\sigma$.

\ding{172} The gain ratio $g$ is used to adjust the overall intensity of the image $\mathbf{I}_1$. Here $g$ initialized to 1 under normal light and over-exposure conditions. In low-light scenarios, $g$ is initialized to 5.

\ding{173} The major axis radius $r_1$ is initialized as 3, and we predict the bias of the variation of $r_1$, then add it to $r_1$.

\ding{174} The minor axis radius $r_2$ is initialized as 2, and we predict the bias of the variation of $r_2$, then add it to $r_2$.

\ding{175} The sharpness filter parameter $\sigma$ is constrained by a Sigmoid activation function to ensure its range is within (0, 1).

The matrix predictor $\mathbb{P_M}$ is responsible for predicting \ding{176} a white balance related parameter $\rho$ and \ding{177} white balance matrix  $\mathbf{E}_{ccm}$ (9 parameters). In total, 10 parameters need to be predicted.

\ding{176} $\rho$ is a hyperparameter of the Minkowski distance in SOG~\cite{Shades_of_gray} white balance algorithm. We set its minimum value to 1 and then use a ReLU activation function followed by adding 1 to restrict its range to (1, +$\infty$).

\ding{177} The matrix $\mathbf{E}_{ccm}$ 
  consists of the 9 parameters predicted by $\mathbb{P_M}$
  and forms a 3x3 matrix. No activation function needs to be added, it would directly added to the identity matrix $\mathbf{E}_{3}$ to form the final $\mathbf{E}_{ccm}$.

\begin{table}[t]
\caption{Ablation analyze on neural implicit 3D LUT $\mathbb{L}$'s dims, we shoe the efficiency comparison ($\#$ Para and Flops), along with mAP comparison on ADE 20K RAW dataset. Flops are calculated from a tensor of size (1, 3, 512, 512).}
\label{tab:ablation_LUT}
\centering
\renewcommand\arraystretch{1.3}
\resizebox{10.8cm}{!}{%

\begin{tabular}{c|cc|ccc}
\toprule
\toprule
$\mathbb{L}$  & \ \# Para $\downarrow$ \ & \ Flops $\downarrow$ \ & mIOU (normal) & mIOU (over-exp) & \ mIOU (dark) \ \\ \hline
dim=16  &  0.93K  &  0.207G &    47.37 ($-$0.58)    &  46.51 ($-$0.11) & 38.16 ($-$0.59) \\ \hline
dim=32  &  1.97K  &  0.784G &    47.95 ($+$0.00) &  46.62 ($+$0.00)  &  38.75 ($+$0.00)  \\ \hline
dim=64  &  12.9K  &  3.041G &  48.02 ($+$0.07)  &     46.72 ($+$0.10)     &  38.75 ($+$0.00) \\ \hline
dim=128 &  50.4K &  11.98G &   47.97 ($+$0.02)   &    46.63 ($-$0.02)    &  38.70 ($-$0.05) \\ 
\bottomrule
\bottomrule
\end{tabular}}
\end{table}

For the neural implicit 3D LUT (NILUT)~\cite{conde2024nilut} $\mathbb{L}$, in the main text, we set the MLP dimension of the neural implicit 3D LUT $\mathbb{L}$ to 32 to save FLOPs, here we test the effects of different dims of $\mathbb{L}$ on the final results, as shown in Table~\ref{tab:ablation_LUT}. Compared to the MLP dimension of 32 in the main text, we observed that setting the MLP dimension of NILUT~\cite{conde2024nilut} to 16 leads to a decrease in performance. Increasing the LUT dimension to 64 results in a slight improvement in performance, but further increasing it to 128 does not lead to performance enhancement. Additionally, as the MLP dimension increases, both the parameter number and FLOPs of $\mathbb{L}$ increase substantially. Therefore, in the experiments of RAW-Adapter, choosing a dimension of 32 for $\mathbb{L}$ is a more reasonable option.




\section{Segmentation on Real-World Dataset~\cite{RAW_segment_dataset}}

\begin{figure}
    \centering
    \includegraphics[width=0.9\linewidth]{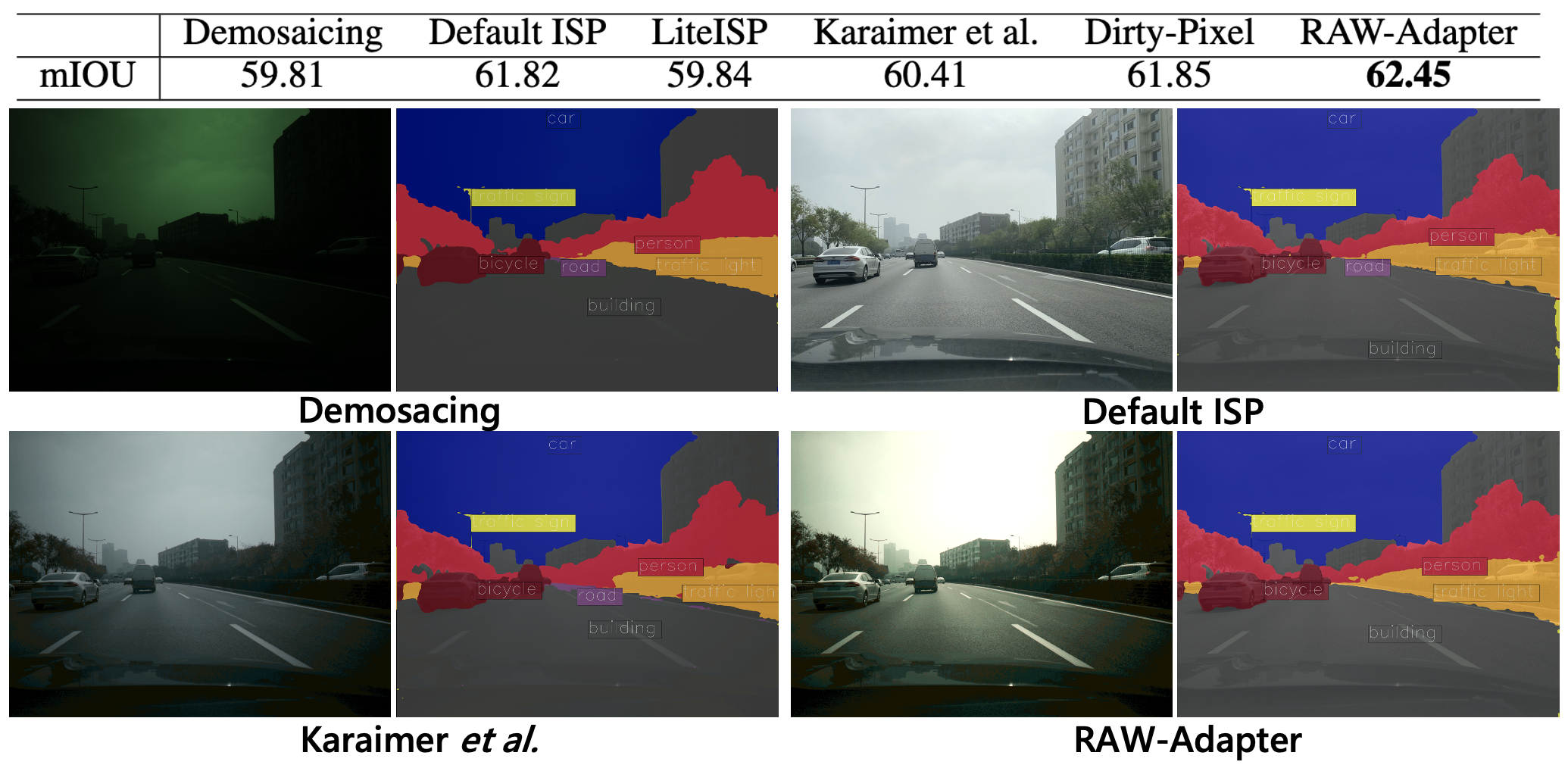}
    \caption{Semantic segmentation results on iPhone XSmax~\cite{RAW_segment_dataset} dataset.}
    \label{fig:iphone_xsmax}
\end{figure}

Additionally, we made the experiments on real-world RAW semantic segmentation dataset iPhone XSmax~\cite{RAW_segment_dataset}, iPhone XSmax consist of 1153 RAW images with their corresponding semnatic labels, where 806 images are set as training set and the other 347 images are set as evaluation set. We adopt Segformer~\cite{xie2021segformer} framework with MIT-B5 backbone, training iters are set to 20000 and other settings are same as ADE 20K RAW's setting. The experimental results are shown in Fig.~\ref{fig:iphone_xsmax}. RAW-Adapter method could also achieve satisfactory results.

\section{Limitation of Current Design}

\textbf{Input-level Adapter} still adopts simple kernel-based denoising and sharpening methods,  this approach is considered for saving computational costs and for simplicity in design, however, we believe that perhaps more advanced denoising methods could bring about better improvements. Another part is that the implicit 3D LUT~\cite{conde2024nilut} is not designed to be image-adaptive, instead, it is a fixed LUT learned from the same dataset, we believe that perhaps an image-adaptive LUT could lead to better improvements, as different images within the same dataset can still have significant variations in information and lighting conditions.

\textbf{Model-level Adapter}'s integration method is still relatively simple. We have extracted intermediate images from the ISP process (I1, I2, I3, I4) to serve as guidance information for designing the model-level adapter. We use the convolution process to simply fuse I1, I2, I3, I4 together to assist the network backbone. We believe that perhaps more effective feature fusion method~\cite{Dong_2024_WACV} could better help improve the performance of downstream tasks.

\section{More Visualization Results}

\begin{figure*}[t]
    \centering
    \includegraphics[width=1.00\linewidth]{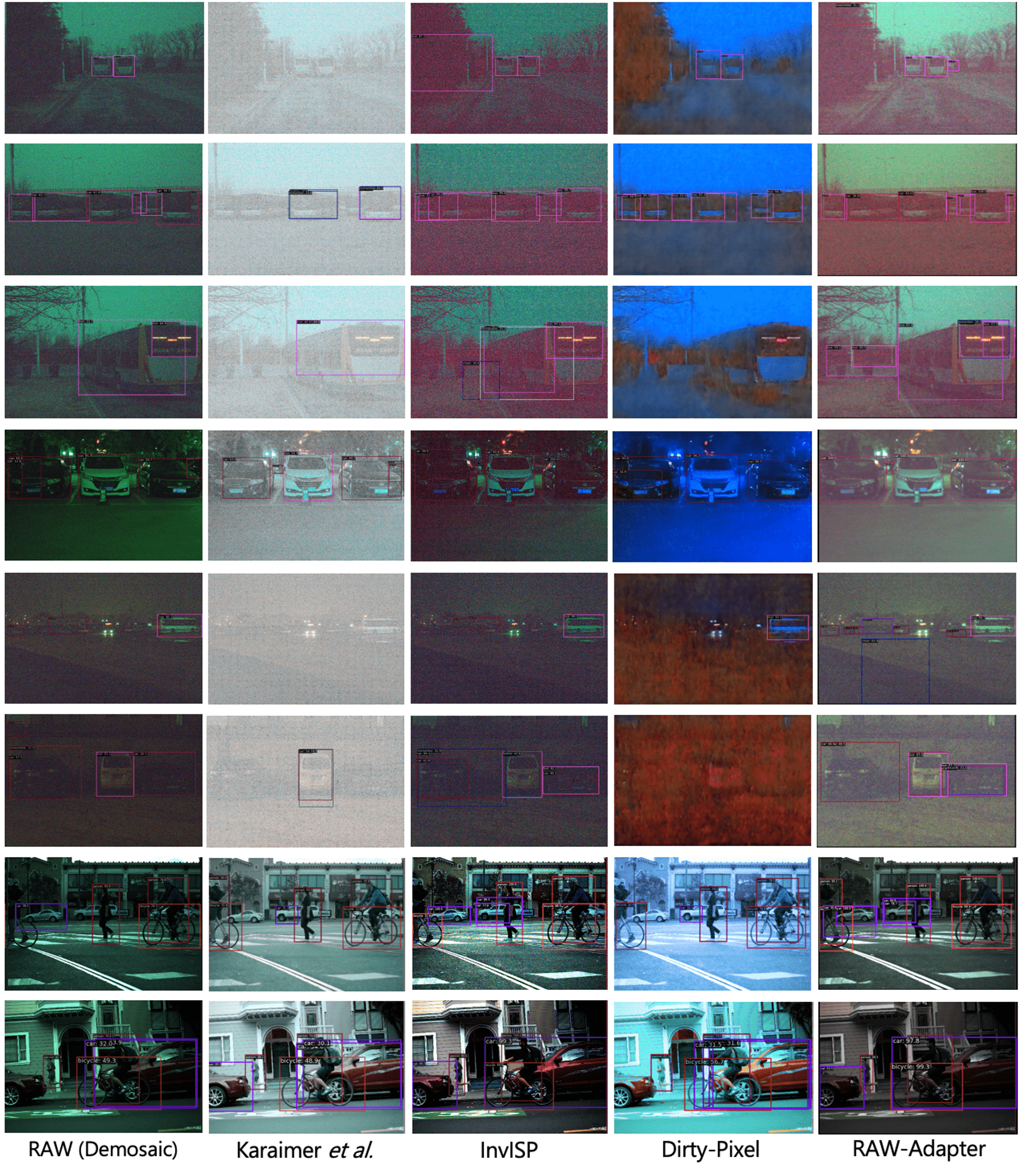}
    \caption{Object detection results on LOD~\cite{LOD_BMVC2021} (line 1 $\sim$ 6) and PASCAL RAW~\cite{omid2014pascalraw} (line 7 and line  8), please zoom in to see details.}
    \label{fig:detection_supp}
    
\end{figure*}

\begin{figure*}[t]
    \centering
    \includegraphics[width=1.00\linewidth]{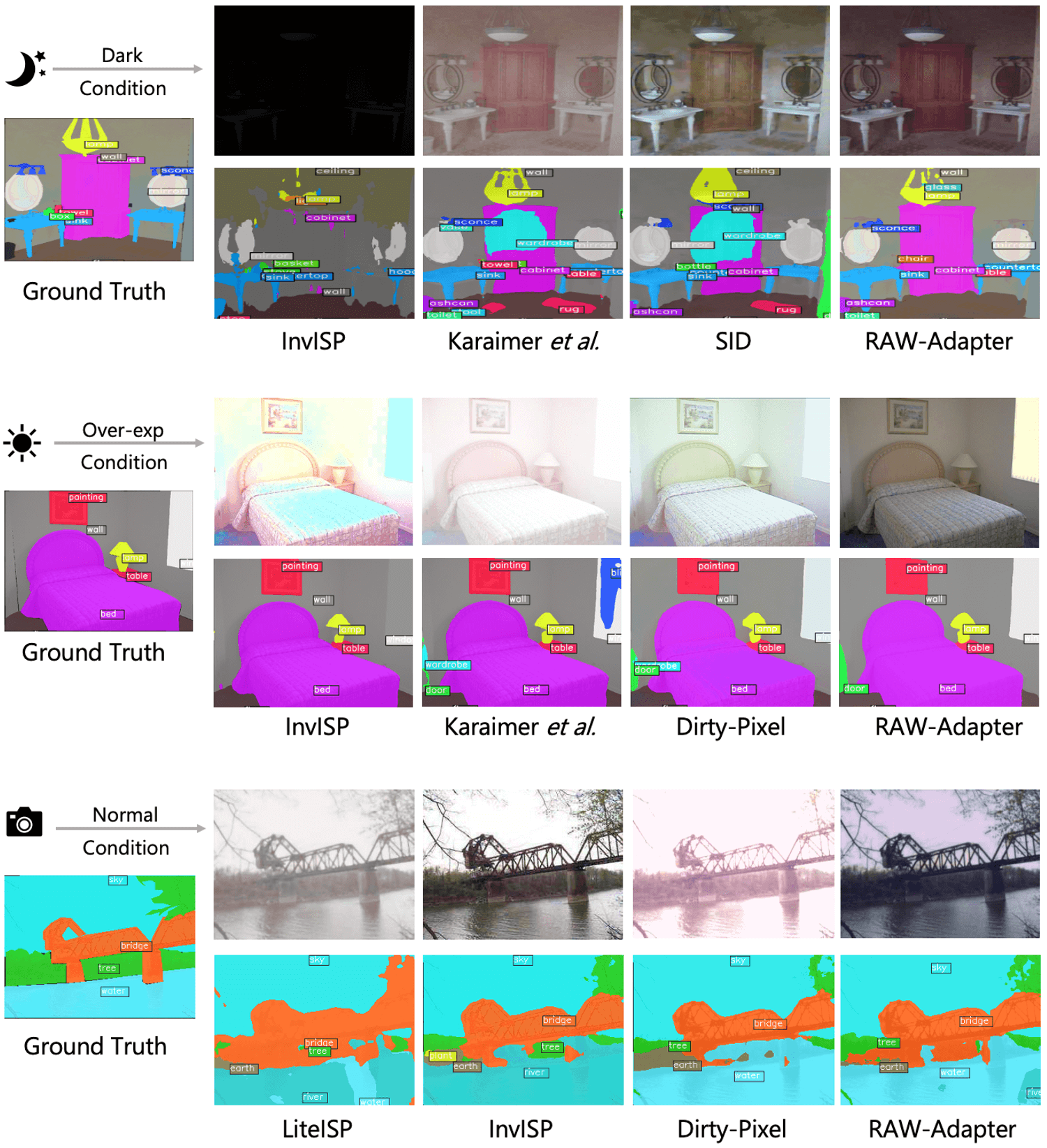}
    \caption{Segmentation results on ADE20K RAW dataset, including dark scene, over-exposure scene and normal scene.}
    \label{fig:segmentation_supp}
    
\end{figure*}

We show more visualization results in this section. The detection results are shown in Fig.~\ref{fig:detection_supp}, where line 1 $\sim$ 6 are the detection results on LOD~\cite{LOD_BMVC2021} dataset and line 7 $\sim$ 8 are the detection results on PASCAL RAW~\cite{omid2014pascalraw} dataset, we show the comparison with ISP methods Karaimer \textit{et al.}~\cite{Michael_eccv16} and InvISP~\cite{invertible_ISP}, along with joint-training method Dirty-Pixel~\cite{steven:dirtypixels2021}. The segmentation results are shown in Fig.~\ref{fig:segmentation_supp}, with comparison of various methods~\cite{steven:dirtypixels2021,Michael_eccv16,SID,RAW-to-sRGB_ICCV2021,invertible_ISP}.

\end{document}